\documentclass[10pt,twocolumn,letterpaper]{article}

\usepackage{wacv}              % To produce the CAMERA-READY version
\usepackage{graphicx}
\usepackage{amsmath}
\usepackage{amssymb}
\usepackage{booktabs}

\usepackage{times}
\usepackage{epsfig}
\usepackage{subcaption}

\usepackage{arydshln}

\usepackage[pagebackref,breaklinks,colorlinks]{hyperref}

\usepackage[capitalize]{cleveref}
\crefname{section}{Sec.}{Secs.}
\Crefname{section}{Section}{Sections}
\Crefname{table}{Table}{Tables}
\crefname{table}{Tab.}{Tabs.}

\def\wacvPaperID{2126} % *** Enter the WACV Paper ID here
\def\confName{WACV}
\def\confYear{2025}

\DeclareMathOperator*{\argmin}{arg\,min}

\begin{document}

\title{Metric Compatible Training for Online Backfilling in Large-Scale Retrieval}

% \and
\author{
Seonguk Seo\textsuperscript{1,3}~~~~Mustafa Gokhan Uzunbas$^{3}$~~~~Bohyung Han$^{1,2}$~~~~Sara Cao$^{3}$  
% \qquad Joena Zhang$^{3}$  \qquad Taipeng Tian$^{3}$
~~~~Ser-Nam Lim$^{4}$  \vspace{0.2cm} \\
 $^{1}$ECE \& $^{2}$IPAI, Seoul National University~~~~$^3$Meta~~~~$^4$University of Central Florida\\
  {\tt\small \{seonguk,gokhanuzunbas,xuefeicao01\}@meta.com},~~{\tt\small bhhan@snu.ac.kr},~~{\tt\small sernam@ucf.edu}
}
\maketitle

% !TEX root = ./../main.tex

%%%%%%%%% ABSTRACT
\begin{abstract}
Backfilling is the process of re-extracting all gallery embeddings from upgraded models in image retrieval systems.
It inevitably spends a prohibitively large amount of computational cost and even entails the downtime of the service.
Although backward-compatible learning sidesteps this challenge by tackling query-side representations, this leads to suboptimal solutions in principle because gallery embeddings cannot benefit from model upgrades.
We address this dilemma by introducing an online backfilling algorithm, which enables us to achieve a progressive performance improvement during the backfilling process without sacrificing the full performance of the new model after the completion of backfilling.
To this end, we first show that a simple distance rank merge is a reasonable option for online backfilling.
Then, we incorporate a reverse transformation module for more effective and efficient merging, which is further enhanced by adopting metric-compatible contrastive learning.
These two components help to make the distances of old and new models compatible, resulting in desirable merge results during backfilling with no extra computational overhead.
Extensive experiments show the benefit of our framework on four standard benchmarks in various settings.

\iffalse
Model upgrades in image retrieval inevitably involve time-consuming backfilling process that re-extracts all gallery embeddings from the upgraded model.
%because the new embedding space is not compatible to previous one.
Backward compatible learning has been investigated to overcome this bottleneck by learning compatible representations, but this entails an inherent trade-off between feature compatibility and discriminability, which may not be overcome.
To address this dilemma, we suggest a practical relaxation, on-the-fly backfilling, which aims to leverage gradual performance improvement as backfill progresses.
We first propose a distance-based retrieval merging approach that always enables on-the-fly backfilling in any model upgrade scenario, which does not sacrifice the performance of final model.
Based on this approach, for more effective and efficient merge, we propose metric compatible contrastive learning with reverse transformation module.
This helps to make the distance metrics of old and new models compatible to each other, resulting in stronger merge results during backfilling without computational overhead.
The extensive experiments demonstrate the effectiveness of our framework on four standard benchmarks in various settings.
\fi
\end{abstract}
%\blfootnote{\textsuperscript{$\dagger$} This work was mostly done during an internship at Meta AI.}
%----------------------------
%          Introducton
% ----------------------------
% !TEX root = ./../main.tex

\section{Introduction}
\label{sec:introduction}

Image retrieval models~\cite{sun2014deep, li2014deepreid, gordo2016deep, wang2018devil} have achieved remarkable performance by adopting deep neural networks for representing images.
Yet, all models need to be upgraded at times to take advantage of improvements in training datasets, network architectures, and training techniques.
This unavoidably leads to the need for re-extracting the features from millions or even billions of gallery images using the upgraded new model.
This process, called \textit{backfilling} or \textit{re-indexing}, needs to be completed before the retrieval system can benefit from the new model, which may take months in practice.

To sidestep this bottleneck, several backfilling-free approaches based on backward-compatible learning~\cite{bct, lce, bict, wan2022continual, duggal2021compatibility} have been proposed.
They learn a new model while ensuring that its feature space is still compatible with the old one, thus avoiding the need for updating old gallery features.
Although these approaches have achieved substantial performance gains without backfilling, they achieve feature compatibility at the expense of feature discriminability and their performance is suboptimal.
We argue that backward-compatible learning is not a fundamental solution and backfilling is still essential to accomplish state-of-the-art performance without performance sacrifices.

To resolve this compatibility-discriminability dilemma, we relax the backfill-free constraint and propose a novel online backfilling algorithm equipped with three technical components.
We posit that an online backfilling technique needs to satisfy three essential conditions: 1) immediate deployment after the completion of model upgrade, 2) progressive and non-trivial performance gains in the middle of backfilling, and 3) no degradation of final performance compared to offline backfilling.
To this end, we first propose a fundamental distance rank merge framework to make online backfilling feasible, which retrieves images from both the old and new galleries separately and merge their results to obtain the final retrieval outputs even when backfilling is still ongoing.
While this approach provides a monotonic performance increase with the progress of backfilling regardless of the gallery of interest and network architectures, it requires feature computations twice, once from the old model and another from the new one at the inference stage of a query.
To overcome this limitation, we introduce a reverse transformation module, which is a lightweight mapping network between the old and new embeddings.
The reverse transformation module allows us to obtain the query representations compatible with both the old and new galleries using only a single feature extraction.
On the other hand, however, we notice that the scales of distance in the embedding spaces of the two models could be significantly different.
We resolve the limitation with a metric compatible learning technique, which calibrates the distances of two models via contrastive learning, further enhancing performance of rank merge.
The main contributions of our work are summarized as follows.

\begin{itemize} 
\item We propose an online backfilling approach, a fundamental solution for model upgrades in image retrieval systems, based on distance rank merge to overcome the compatibility-discriminability dilemma in existing compatible learning methods.
\item We incorporate a reverse query transform module to make it compatible with both the old and new galleries while computing the feature extraction of query only once in the middle of the backfilling process.
\item We adopt a metric-compatible learning technique to make the merge process robust by calibrating distances in the feature embedding spaces given by the old and new models.
\item The proposed approach outperforms all existing methods by significant margins on four standard benchmark datasets under various scenarios.
\end{itemize}

\begin{figure*}
	\begin{center}
        \includegraphics[width=0.95\linewidth]{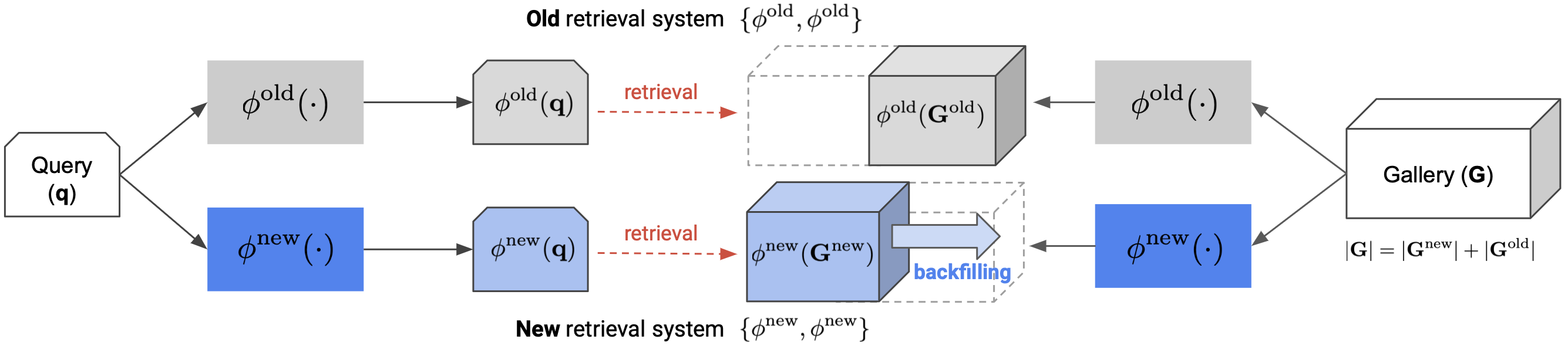}
        \vspace{-0.5cm}
	\end{center}
	\caption{Image retrieval with the proposed distance rank merge technique.
	In the middle of backfilling, we retrieve images  independently using two separate models and their galleries, and merge the retrieval results based on their distances.
	Note that the total number of gallery embeddings are fixed throughout the backfilling process, \ie, $ |\mathbf{G}| = |\mathbf{G}^\text{new}|+|\mathbf{G}^\text{old}|$.
	}
	\vspace{-0.2cm}    
\label{fig:ensemble}
\end{figure*}
%
%The rest of this paper is organized as follows.
%Section~\ref{sec:related} reviews the related works.
%We present the main framework of online backfilling in Section~\ref{sec:method}, and discuss the technical components for improvement in Section~\ref{sec:mct}.
%% Section~\ref{sec:rtm} and \ref{sec:metric_learning}.
%We demonstrate the effectiveness of the proposed framework in Section~\ref{sec:experiments} and conclude this paper in Section~\ref{sec:conclusion}.

%----------------------------
%          related work
% ----------------------------

% !TEX root = ./../main.tex

\section{Related Work}
\label{sec:related}

{
\paragraph{Backward compatible learning}
Backward compatibility refers to the property to support older versions in hardware or software systems.
It has been recently used in model upgrade scenarios in image retrieval systems.
Since the feature spaces given by the models relying on training datasets in different regimes are not compatible~\cite{wang2018towards, li2015convergent}, model upgrades require re-extraction of all gallery images from new models, which takes a huge amount of computational cost.
To prevent this time-consuming backfilling cost, backward compatible training (BCT)~\cite{bct, lce, bai2021dual, pan2022dynamic, wan2022continual, zhang2022towards} has been proposed to learn better feature representations while being compatible with old embeddings, which makes the new model backfill-free.
Shen~\etal~\cite{bct} employ the influence loss that utilizes the old classifier as a regularizer when training the new model.
LCE~\cite{lce} introduces an alignment loss to align the class centers between old and new models and a boundary loss that restricts more compact intra-class distributions for the new model.
Bai~\etal~\cite{bai2021dual} propose a joint prototype transfer with structural regularization to align two embedding features.
UniBCT~\cite{zhang2022towards} presents a structural prototype refinement algorithm that first refines noisy old features with graph transition and then conducts backward compatible training.
Although these approaches improved compatible performance without backfilling, they clearly sacrifice feature discriminability to achieve feature compatibility with non-ideal old gallery embeddings.
}

\vspace{-2mm}
\paragraph{Compatible learning with backfilling}
To overcome the inherent limitation of backward compatible learning, several approaches~\cite{fct, ract, bict} have been proposed to utilize backfilling but efficiently.
Forward compatible training (FCT)~\cite{fct} learn a lightweight transformation module that updates old gallery embeddings to be compatible with new embeddings.
Although it gives better compatible performance than BCT, it requires an additional side-information~\cite{chen2020simple} to map from old to new embeddings, which limits its practicality.
Moreover, FCT still suffers from computational bottleneck until all old gallery embeddings are transformed, especially when the side-information needs to be extracted.
On the other hand, RACT~\cite{ract} and BiCT~\cite{bict} alleviate this bottleneck issue by backfilling the gallery embeddings in an online manner.
RACT first trains a backward-compatible new model with regression-alleviating loss, then backfills the old gallery embeddings with the new model.
Because the new feature space is compatible with the old one, the new model can be deployed right away while backfilling is carried out in the background.
BiCT further reduces the backfilling cost by transforming the old gallery embeddings with forward-compatible training~\cite{fct}.
Although both approaches can utilize online backfilling, they still sacrifice the final performance because the final new embeddings are constrained by the old ones.
Unlike these methods, our framework enables online backfilling while fully exploiting the final new model performance without any degradation.

%%%%%%%%%%%%%%%%%%%%%%%%%%%%%%%%%%%%%%%%%%%%%%%%%%%%%%%%%%%%%%%%%%%%%%%%%%%%%%%%%%%%%%

%----------------------------
%          Method
% ----------------------------
% !TEX root = ./../main.tex

\section{Image Retrieval by Rank Merge}
\label{sec:method}

This section discusses our baseline image retrieval algorithm that makes online backfilling feasible. 
We first present our motivation and then describe technical details with empirical observations.

\subsection{Overview}
\label{sec:overview}
Our goal is to develop a fundamental solution to overcome the compatibility-discriminability trade-off via online backfilling in compatible model upgrade.
We aim to remove inherent limitations of backfill-free backward-compatible learning---the inability to benefit from upgraded representations of gallery images extracted from new models---while avoiding system downtime and outdated service during offline backfilling process until the completion of backfilling.
% prohibitive computational cost, including the situation that we cannot benefit from model upgrade of the offline backfilling process, until backfilling is completed.

Image retrieval systems with online backfilling should satisfy the following three conditions: 
\begin{enumerate}
    \item The system can be deployed as soon as the model upgrade is completed. \vspace{-1mm}
    \item The performance should monotonically increase without negative flips\footnote{The ``negative flip" refers to performance degradation caused by incorrect retrievals in the new model, which were correct in the old model.} as backfill progresses. \vspace{-1mm}
    \item The final performance should be better than or equivalent to the algorithm relying on offline backfilling.
\end{enumerate}
We present a distance rank merge approach for image retrieval, which enables online backfilling in arbitrary model upgrade scenarios.
This method maintains two separate retrieval pipelines corresponding to the old and new models and merges the retrieval results from the two models based on distances from a query embedding.
This allows us to run the retrieval system without a warm-up period and achieve surprisingly good results during the backfilling process.
Note that the old and new models are not required to be compatible at this moment but we will make them so to further improve performance in the subsequent sections.

%%%%%%%%%%%  MERGING RESULTS  %%%%%%%%%%%
% !TEX root = ./../../main.tex

\begin{figure*}[]
    \centering
    \begin{subfigure}[b]{0.475\textwidth}   
        \centering 
        \includegraphics[width=\textwidth]{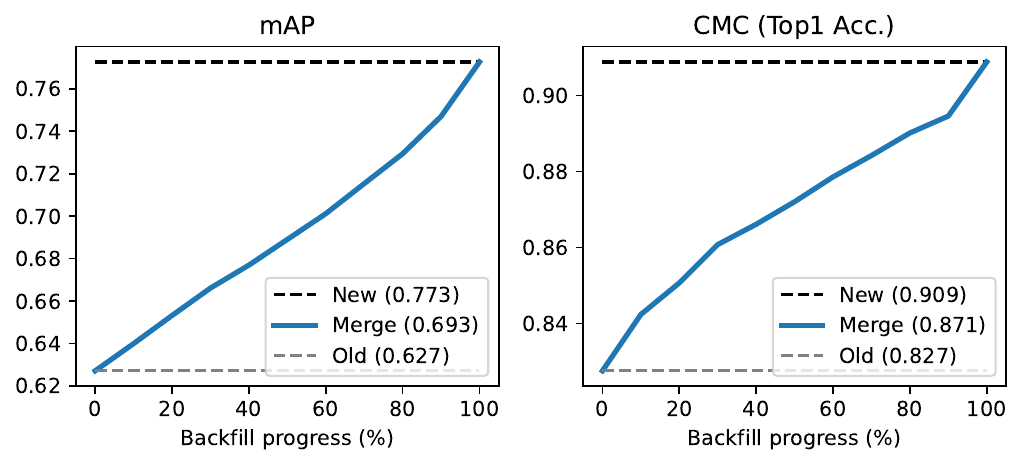}
        \caption[]%
        {{\small ImageNet-1K}}    
        \label{fig:mean and std of net44}
    \end{subfigure}
    \hfill
%    \begin{subfigure}[b]{0.475\textwidth}
%        \centering
%        \includegraphics[width=\textwidth]{figures/exp/ens/ens_cifar2.pdf}
%        \caption[Network2]%
%        {{\small CIFAR-100}}    
%        \label{fig:mean and std of net14}
%    \end{subfigure}
    \begin{subfigure}[b]{0.475\textwidth}   
        \centering 
        \includegraphics[width=\textwidth]{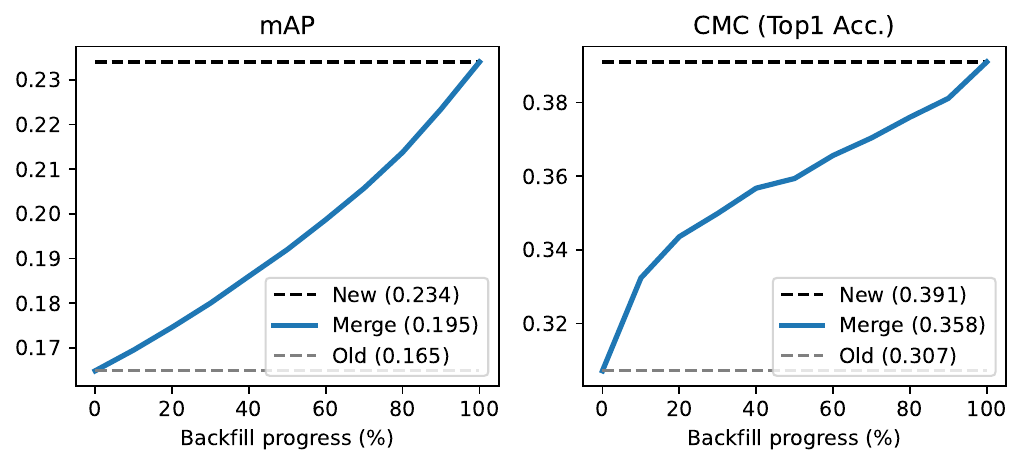}
        \caption[]%
        {{\small Places-365}}    
        \label{fig:mean and std of net34}
    \end{subfigure}
%    \hfill
%    \begin{subfigure}[b]{0.475\textwidth}  
%        \centering 
%        \includegraphics[width=\textwidth]{figures/exp/ens/ens_market2.pdf}
%        \caption[]%
%        {{\small Market-1501}}    
%        \label{fig:mean and std of net24}
%    \end{subfigure}
\vspace{-2mm}
    \caption[ The average and standard deviation of critical parameters ]
    {\small mAP and CMC results on the standard benchmarks using ResNet-18.
    \textit{Old} and \textit{New} denote the performance without backfilling and with offline backfilling, respectively.
    The distance rank merge of the old and new models, denoted by \textit{Merge}, exhibits desirable results; the accuracy monotonically increases as backfill progresses without negative flips for all datasets and the rank merge with online backfilling achieves competitive final performances of offline backfilling.
    The numbers in the legend indicate either AUC$_\text{mAP}$ or AUC$_\text{CMC}$ scores.
    } 
    \label{fig:ensemble_results}
    \vspace{-2mm}
\end{figure*}

\subsection{Formulation}
\label{sec:formulation}

Let $\mathbf{q} \in \mathbf{Q}$ be a query image and $\mathbf{G}=\{\mathbf{g}_1, ..., \mathbf{g}_N\}$ be a gallery composed of $N$ images.
An embedding network $\phi(\cdot)$ projects an image onto a learned feature space.
To retrieve the closest gallery image given a query, we solve $\argmin_{\mathbf{g} \in \mathbf{G}} {\text{dist}}\left(\phi(\mathbf{q}), \phi(\mathbf{g})\right)$, where ${\text{dist}}(\cdot, \cdot)$ is a distance metric.
Following~\cite{bct}, the retrieval accuracy is given by
\begin{align}
    \mathcal{M}(\phi(\mathbf{Q}), \phi(\mathbf{G})),    
    \label{eq:performance}
\end{align}
where $\mathcal{M}(\cdot, \cdot)$ is an evaluation metric such as mean average precision (mAP) or cumulative matching characteristics (CMC).
% and $\phi(\cdot)$ indicates embedding models for query or gallery.
%
\iffalse
Following~\cite{bct}, we define the retrieval performance as
\begin{align}
    \mathcal{M}(\phi_{\theta_\text{q}}(\mathbf{Q}), \phi_{\theta_\text{g}}(\mathbf{G})),    
    \label{eq:performance}
\end{align}
where $\mathcal{M}(\cdot, \cdot)$ is an evaluation metric such as mean average precision (mAP) and cumulative matching characteristics (CMC), and $\phi_{\theta_\text{q}}(\cdot)$ and $\phi_{\theta_\text{g}}(\cdot)$ indicate embedding models for query and gallery, respectively.
We refer to~\eqref{eq:performance} as self-test performance when $\phi_{\theta_\text{q}} = \phi_{\theta_\text{g}}$ and cross-test performance, otherwise.
\fi

\vspace{-2mm}
\paragraph{Backward compatibility} 
Denote the old and new embedding networks by $\phi^\text{old}(\cdot)$ and $\phi^\text{new}(\cdot)$ respectively.
If $\phi^\text{new}(\cdot)$ is backward compatible with $\phi^\text{old}(\cdot)$, then we can perform search on a set of old gallery embeddings using a new query embedding, \ie, $\argmin_{\mathbf{g} \in \mathbf{G}} {\text{dist}}(\phi^\text{new}(\mathbf{q}), \phi^\text{old}(\mathbf{g}))$.
As stated in~\cite{bct}, the backward compatibility is achieved when the following criterion is satisfied:
\begin{align}
    \mathcal{M}(\phi^\text{new}(\mathbf{Q}), \phi^\text{old}(\mathbf{G})) > \mathcal{M}(\phi^\text{old}(\mathbf{Q}), \phi^\text{old}(\mathbf{G})).
\end{align}
From now, we refer to a pair of embedding networks for query and gallery as a retrieval system, \eg, $\{ \phi^{(\cdot)}, \phi^{(\cdot)} \}$.

\vspace{-2mm}
\paragraph{Rank merge} 
Assume that the first $M$ out of a total of $N$ images are backfilled, \textit{i.e.}, $\mathbf{G}^\text{new} = \{\mathbf{g}_1, ..., \mathbf{g}_M\}$ and $\mathbf{G}^\text{old} = \{\mathbf{g}_{M+1}, ..., \mathbf{g}_N\}$.
Note that the total number of stored gallery embeddings is fixed to $N$ during the backfilling process, \ie, $\mathbf{G}^\text{old} = \mathbf{G} - \mathbf{G}^\text{new}$.
Then, we first conduct image retrieval using the individual retrieval systems, $\{ \phi^\text{old}, \phi^\text{old} \}$ and $\{ \phi^\text{new}, \phi^\text{new} \}$, independently as
\begin{align}
\mathbf{g}_m &= \argmin_{\mathbf{g}_i \in \mathbf{G}^\text{old}}~\text{dist} \left(\phi^\text{old}(\mathbf{q}), \phi^\text{old}(\mathbf{g}_i) \right), \\
\mathbf{g}_n &= \argmin_{\mathbf{g}_j \in \mathbf{G}^\text{new}}~\text{dist} \left(\phi^\text{new}(\mathbf{q}), \phi^\text{new}(\mathbf{g}_j) \right).
\end{align}
Figure~\ref{fig:ensemble} illustrates the retrieval process.
For each query image $\mathbf{q}$, we finally select $\mathbf{g}_m$ if $\text{dist} (\phi^\text{old}(\mathbf{q}), \phi^\text{old}(\mathbf{g}_m) ) < \text{dist}(\phi^\text{new}(\mathbf{q}), \phi^\text{new}(\mathbf{g}_n))$ and $\mathbf{g}_n$ otherwise.
The retrieval performance after rank merge during backfilling is given by
\begin{align}
   \mathcal{M}_t & := \label{eq:rankmerge_metric} \\
    & \mathcal{M}(\{ \phi^\text{old}(\mathbf{Q}), \phi^\text{new}(\mathbf{Q})\}, \{ \phi^\text{old}(\mathbf{G}_t^\text{old}), \phi^\text{new}(\mathbf{G}_t^\text{new}) \}), \nonumber
\end{align}
where $t \in [0, 1]$ indicates the rate of backfilling completion, \ie, $|\mathbf{G}_t^\text{new}| = t|\mathbf{G}|$ and $|\mathbf{G}_t^\text{old}| = (1-t)|\mathbf{G}|$.
The criteria discussed in Section~\ref{sec:overview} are formally defined as
\begin{align}
    &\mathcal{M}_0 \geq \mathcal{M}(\phi^\text{old}(\mathbf{Q}), \phi^\text{old}(\mathbf{G})), \label{eq:ens_0} \\
    &\mathcal{M}_1 \geq \mathcal{M}(\phi^\text{new}(\mathbf{Q}), \phi^\text{new}(\mathbf{G})), \label{eq:ens_1} \\
    &\mathcal{M}_{t_1} \geq \mathcal{M}_{t_2}~~\text{if}~~t_1 \geq t_2.
    \label{eq:ens_}
\end{align}

\vspace{-2mm}
\paragraph{Comprehensive evaluation}
To measure both backfilling cost and model performance comprehensively during online backfilling, we utilize the following metrics that calculate the area under mAP or CMC curves as
\begin{align}
    \text{AUC}_\text{mAP} = \hspace{-1mm} \int_{0}^1 \text{mAP}_t dt ~~\text{and}~~ %\label{eq:aumap}\\
    \text{AUC}_\text{CMC} = \hspace{-1mm} \int_{0}^1 \text{CMC}_t dt. \nonumber %\label{eq:aucmc}
\end{align}

\subsection{Merge Results}

We present the results from the rank merge strategy on two standard benchmarks, including ImageNet-1K~\cite{imagenet} and Places-365~\cite{places365}, in Figure~\ref{fig:ensemble_results}.
Despite its simplicity, the rank merging approach yields strong and robust results for all datasets; both mAP and CMC monotonically increase without negative flips as backfill progresses even though the old and new models are not compatible with each other. 
%Also, it takes full advantage of the new model {until} the end of backfilling without suffering from performance degradation.
Unlike other backward-compatible learning or online backfilling methods~\cite{bct, ract, bict}, this method takes full advantage of the new model at the end of backfilling without witnessing performance degradation.
This validates that the rank merge technique satisfies the criteria for online backfilling discussed in Section~\ref{sec:overview} and~\ref{sec:formulation}.
Please refer to Section~\ref{sec:setting} for detailed experimental results.

%\input{./sections/Method_reverse_module}
%\input{./sections/Method_metric_learning}
% !TEX root = ./../main.tex

%\clearpage
\section{Metric Compatible Training}
\label{sec:mct}

Our baseline image retrieval method is model-agnostic, free from extra training, and effective for monotonic performance improvement in online backfilling.
%However, there exists room for improvement in calibrating distance metrics of feature embedding spaces of both systems for effective rank merge.
However, there still exists room for improvement by reducing computational cost at the inference stage and calibrating distance metrics in old and new feature embedding spaces for effective rank merge.
This section discusses the issues in detail and presents how we figure them out.
%However, one may argue that the proposed approach is computationally expensive at inference time because we need to conduct feature extraction twice per query for both the old and new models.

%\begin{figure}
%	\begin{center}
%        \includegraphics[width=0.9\linewidth]{figures/rtm_1.png}
%        \vspace{-3mm}
%	\end{center}
%	\caption{Reverse query transform module, $\psi(\cdot)$, learns a mapping from new to old feature spaces.
%	We only update the parameters of the module $\psi(\cdot)$ (in red rectangle) during training.
%	}
%	\label{fig:rtm}
%	\vspace{-2mm}    
%\end{figure}

\subsection{Reverse Query Transform}
\label{sub:formulation}

\begin{figure*}
	\begin{center}
        \includegraphics[width=0.97\linewidth]{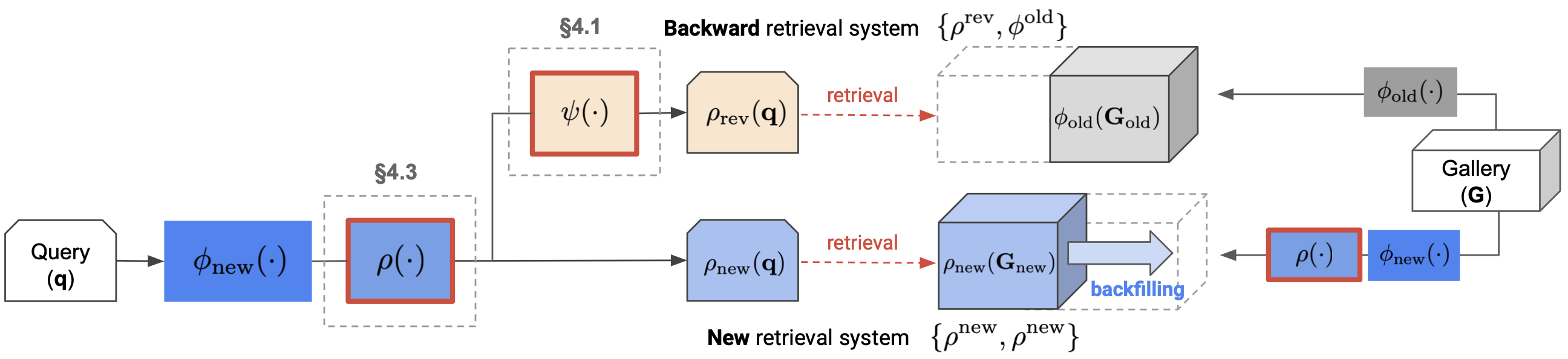}
        \vspace{-5mm}
	\end{center}
	\caption{
%	Image retrieval merging with reverse query transform module.
Image retrieval with our final rank merge framework including Section \ref{sub:formulation}-\ref{sub:learnnew}.
%	Backward retrieval system consists of reversely transformed new query and old gallery, $\{ \phi^\text{rev}, \phi^\text{old} \}$.
%	The final image retrieval results are given by merging the outputs from $\{ \phi^\text{rev}, \phi^\text{old} \}$ and $\{ \phi^\text{new}, \phi^\text{new} \}$.
Backward retrieval system consists of reversely transformed new query and old gallery, $\{ \rho^\text{rev}, \phi^\text{old} \}$.
	The final image retrieval results are given by merging the outputs from $\{ \rho^\text{rev}, \phi^\text{old} \}$ and $\{ \rho^\text{new}, \rho^\text{new} \}$.
	}
	\label{fig:ensemble_rtm}
	\vspace{-1mm}    
\end{figure*}

To reduce the computational cost incurred by computing query embeddings twice for two separate systems at the inference stage, we compute the embedding using the new model and transform it to the version compatible with the old model through the reverse query transform module.
% as illustrated in Figure~\ref{fig:rtm}.
To establish such a mechanism, we fix the parameters of the old and new models $\{ \phi^\text{old}, \phi^\text{new} \}$ after training them independently, and train a lightweight network, $\psi (\cdot)$, which transforms the embedding in the new model to the one in the old model. 
For each training example $\mathbf{x}$, our objective is minimizing the following loss:
\begin{align}
\mathcal{L}_\text{{RQT}}(\mathbf{x}) := \text{dist}\left( \psi \left(\phi^\text{new}(\mathbf{x})\right), \phi^\text{old}(\mathbf{x})
\right),
\label{eq:rev_basic}
\end{align}
where dist$(\cdot, \cdot)$ is a distance metric such as $\ell_2$ or cosine distances.
\iffalse
However, we found that using only this alignment loss makes embedding easily overfit to old space and aggravates feature discriminability.
To avoid overfitting, we employ a classification loss on top of reverse-transformed new embedding as a regularizer.
The final objective is
%
\begin{align}
\min_{\psi_{\text{rev}}} \ell(x, y) &:= \text{dist}\bigg(\phi_\text{old}(x),~\psi_{\text{rev}}\Big(\phi_\text{new}(x)\Big)\bigg) \nonumber \\
&+ \lambda\cdot \text{CE}\bigg(y,~h_\text{new}\Big(\psi_{\text{rev}} \big(\phi_\text{new}(x)\big) \Big)\bigg),
\label{eq:rev}
\end{align}
%
where $y$ is the label of input image $x$, $h_\text{new}(\cdot)$ is the classification head of new model, and \text{CE}$(\cdot, \cdot)$ and $\lambda$ denote cross-entropy loss and its weight hyperparameter.
Note that we freeze the classification head of the new model, $h_\text{new}(\cdot)$, to guide the transformed embedding that satisfies the `rules' of original new embedding, which helps to preserve feature discriminability.
\fi
Because we only update the parameters in $\psi(\cdot)$, not the ones in $\phi^\text{new}(\cdot)$ or $\phi^\text{old}(\cdot)$, we can still access the representations given by the new model at no cost even after the optimization of $\psi(\cdot)$.
Note that this reverse query transform module differs from FCT~\cite{fct} mainly in terms of transformation direction and requirement of side information. FCT performs a transformation from the old representation to the new, while the opposite is true for our proposed approach.
Since the embedding quality of a new model is highly likely to be better than that of an old one, our reverse transformation module performs well even without additional side information and, consequently, is more practical and efficient.%

\vspace{-2mm}
%\subsection{Integration into Baseline Retrieval System}
\paragraph{Integration into baseline retrieval system}

Figure~\ref{fig:ensemble_rtm} illustrates the distance rank merge process together with the proposed reverse transformation module.
The whole procedure consists of two retrieval systems defined by a pair of query and gallery representations, backward retrieval system $\{ \phi^\text{rev}, \phi^\text{old} \}$ and new retrieval system $\{ \phi^\text{new}, \phi^\text{new} \}$, where $\phi^\text{rev} := \psi(\phi^\text{new})$.
Note that we obtain both the new and compatible query embeddings, $\phi^\text{new}(\mathbf{q})$ and $\phi^\text{rev}(\mathbf{q}) = \psi (\phi^\text{new}(\mathbf{q}))$, using a shared feature extractor, $\phi^\text{new}(\cdot)$.

The entire image retrieval pipeline consists of two parts: 1) feature extraction of a query image and 2) search for the nearest image in a gallery from the query.
Compared to the image retrieval based on a single model, the computational cost of the proposed model with rank merge requires negligible additional cost which corresponds to feature transformation $\psi(\cdot)$ in the first part.
%\footnote{Additional computational cost is less than 0.1\% of those of ResNet-18.},
{Note that the number of total gallery embeddings is fixed, \ie, $|\mathbf{G}^\text{new}|+|\mathbf{G}^\text{old}| = |\mathbf{G}|$, so the cost of the second part is almost the same in both cases.}

%\section{Distance Calibration}
%\label{sec:metric_learning}

%\subsection{Motivation}
%While the proposed rank merge technique with the basic reverse transformation module works well, there exists room for improvement in calibrating feature embedding spaces of both systems.
%This section discusses the issues in details and presents how we figure them out.

%\subsection{Cross-Model Contrastive Learning}
\subsection{Metric Compatible Contrastive Learning}
\label{sub:mccl}
While the rank merge with the basic reverse transformation works well, the objective in \eqref{eq:rev_basic} cares about the positive pairs $\phi^\text{old}$ and $\phi^\text{rev}$ with no consideration of negative ones, which sometimes lead to misranked position.
To handle this issue, we employ a supervised contrastive learning loss~\cite{InfoNCE, khosla2020supervised} to consider both positive and negative pairs as
%
%\vspace{-1mm}
\begin{align}
\mathcal{L}_\text{CL} (\mathbf{x}_i, y_i) &= - \log \frac{\sum_{y_k=y_i} s_{ik}^\text{old}}{\sum_{y_k=y_i} s_{ik}^\text{old} + \sum_{y_k \neq y_i} s_{ik}^\text{old}},
\label{eq:cl}
\end{align}
where $s_{ij}^\text{old} = \exp\left(-\text{dist}\left(\phi^\text{rev}(\mathbf{x}_i), \phi^\text{old}(\mathbf{x}_j)\right) \right)$ and $y_i$ denotes the class membership of the $i^\text{th}$ sample.
For more robust contrastive training, we perform hard example mining for both the positive and negative pairs\footnote{For each anchor, we select the half of the examples in each of positive and negative labels based on the distances from the anchor.}.
Such a contrastive learning approach facilitates distance calibration and improves feature discrimination because it promotes separation of the positive and negative examples.

\begin{figure}
	\begin{center}
        \includegraphics[width=0.99\linewidth]{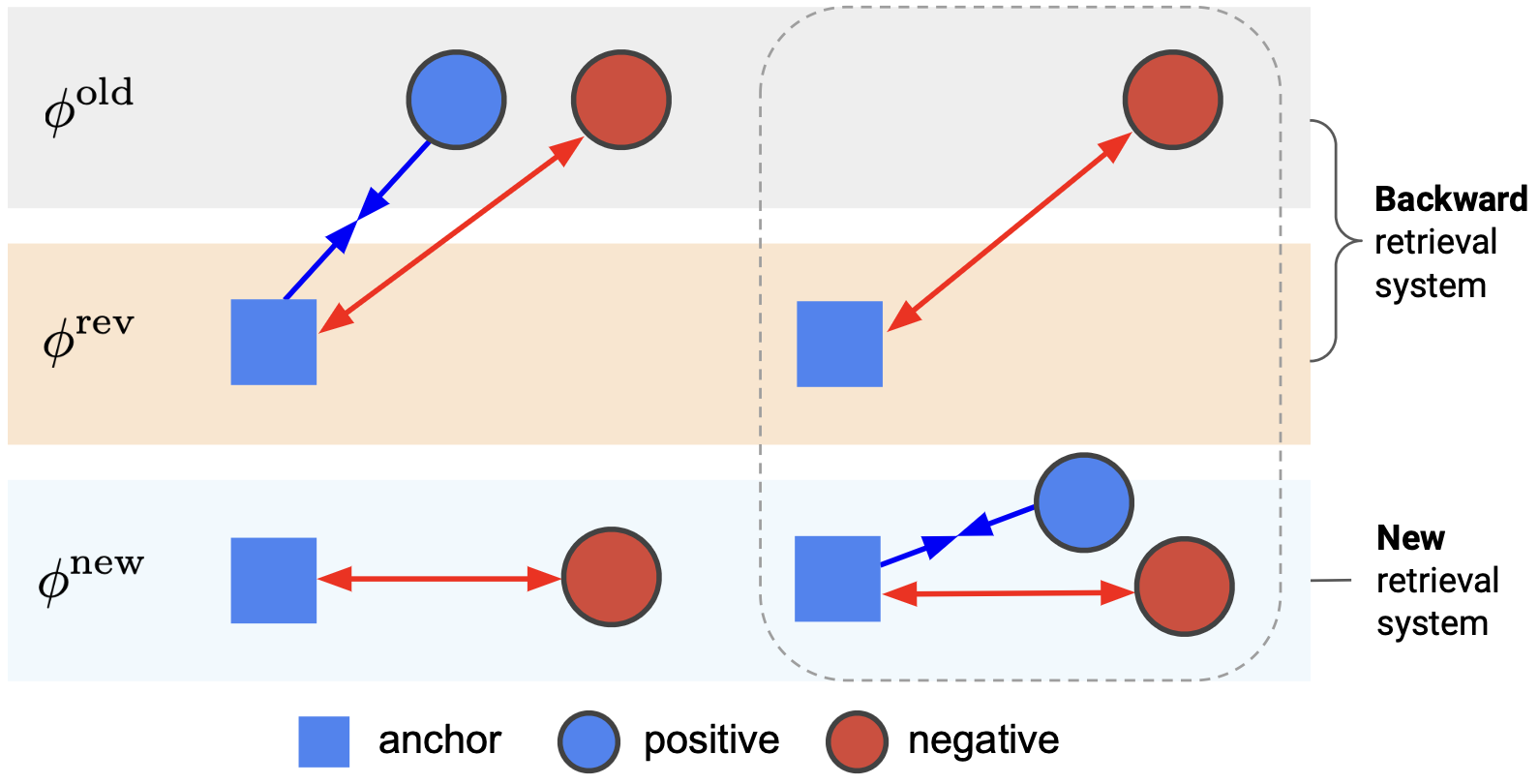}
        \vspace{-0.5cm}
	\end{center}
	\caption{Illustration of metric compatible contrastive learning loss with backward retrieval system $\{ \phi^\text{old}, \phi^\text{rev} \}$ and new retrieval system $\{ \phi^\text{new}, \phi^\text{new} \}$.
	Two boxes with dotted lines corresponds to two terms in~\eqref{eq:cmcl}.
	For each retrieval system, the distances between positive pairs are learned to be both smaller than those of negative pairs in the two systems.
	}
	\label{fig:mccl}
	\vspace{-1mm}    
\end{figure}

Now, although the distances within the backward retrieval system $\{ \phi^\text{rev}, \phi^\text{old} \}$ become more comparable, they are still not properly calibrated in terms of the distances in the new retrieval system $\{ \phi^\text{new}, \phi^\text{new} \}$.
%\suseo{However, \eqref{eq:cl} itself does not make the distances of backward retrieval system $\{ \phi^\text{rev}, \phi^\text{old} \}$ fully comparable with the new retrieval system $\{ \phi^\text{new}, \phi^\text{new} \}$. }
Considering distances in both retrieval systems jointly when we train the reverse transformation module, we can obtain more comparable distances and consequently achieve more reliable rank merge results.
From this perspective, we propose a metric compatible contrastive learning loss as
\begin{align}
\label{eq:cmcl}
\mathcal{L}_\text{MCL}&(\mathbf{x}_i, y_i) =  \\
&- \log \frac{\sum_{y_k=y_i} s^\text{old}_{ik} }{\sum_{y_k=y_i} s^\text{old}_{ik} + \sum_{y_k \neq y_i} s^\text{old}_{ik} + \sum_{y_k\neq y_i} s^\text{new}_{ik}  } \nonumber \\
&-\log \frac{\sum_{y_k=y_i} s^\text{new}_{ik} }{\sum_{y_k=y_i} s^\text{new}_{ik} + \sum_{y_k\neq y_i} s^\text{new}_{ik} + \sum_{y_k\neq y_i} s^\text{old}_{ik}  }, \nonumber 
\end{align}
where $s^\text{new}_{ij} = \exp(-\text{dist}\big(\phi^\text{new}(\mathbf{x}_i), \phi^\text{new}(\mathbf{x}_j)\big))$ and $s^\text{old}_{ij} = \exp(-\text{dist}\big(\phi^\text{rev}(\mathbf{x}_i), \phi^\text{old}(\mathbf{x}_j)\big))$.
Figure~\ref{fig:mccl} depicts the concept of the loss function.
The positive pairs from the backward retrieval system $\{ \phi^\text{rev}, \phi^\text{old} \}$ are trained to locate closer to the anchor than not only the negative pairs from the same system but also the ones from the new system $\{ \phi^\text{new}, \phi^\text{new} \}$, and vice versa.
We finally replace \eqref{eq:rev_basic} with \eqref{eq:cmcl} for training the reverse transformation module.
Compared to~\eqref{eq:cl}, additional heterogeneous negative terms in the denominator of \eqref{eq:cmcl} play a role as a regularizer to make the distances from one model directly comparable to those from other one, which is desirable for our rank merge strategy.

The loss in~\eqref{eq:cmcl} looks similar to the one proposed in~\cite{ract}, however, ours focuses on distance compatibility between old and new feature spaces instead of feature compatibility.
%does not constrain the new model by feature compatibility but focuses only on distance compatibility between two spaces for our rank merge.
%However, while \cite{ract} directly aims at feature compatibility with old space, our method focuses only on metric compatibility between two spaces for rank merge, because our merge framework does not require old and new models to be compatible to each other.
As shown in other backward-compatible methods~\cite{bct, bict}, achieving feature compatibility sacrifices performance of new model due to constrained flexibility to old model; our metric compatible training with enhanced flexibility is free from this issue, resulting in superior merge results.
%directly constrain the feature space of new model 

\subsection{Learnable New Feature Embedding}
\label{sub:learnnew}
\begin{figure}
	\begin{center}
        \includegraphics[width=0.95\linewidth]{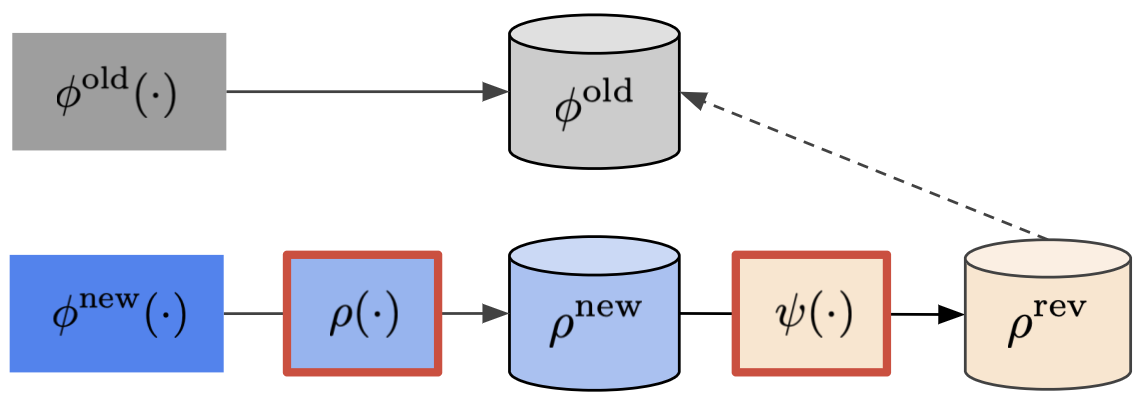}
        \vspace{-5mm}
	\end{center}
	\caption{Compatible training with learnable new embedding, where another transformation module $\rho(\cdot)$ is incorporated on top of the new model to learn new embedding favorable to our rank merging.
%	Compared to Figure~\ref{fig:rtm}, another transformation module $\rho(\cdot)$ is incorporated on top of the new model to learn new embedding favorable to our rank merging. 
	The retrieval results are now merged from $\{ \rho^\text{rev}, \phi^\text{old} \}$ and $\{ \rho^\text{new}, \rho^\text{new} \}$.
	}
	\label{fig:rtm_trainable}
	\vspace{-1mm}    
\end{figure}

Until now, we do not jointly train the reverse transformation module $\psi(\cdot)$ and the new feature extraction module $\phi^\text{new}(\cdot)$.
% as illustrated in  Figure~\ref{fig:rtm}.
This hampers the distance compatibility between the backward and new retrieval systems because the backward retrieval system $\{ \phi^\text{rev}, \phi^\text{old} \}$ is the only part to be optimized while the new system $\{ \phi^\text{new}, \phi^\text{new} \}$ is fixed.
To provide more flexibility, we add another transformation module $\rho(\cdot)$ on top of the new model as shown in Figure~\ref{fig:rtm_trainable}, where  $\rho^\text{new} = \rho(\phi^\text{new})$ and $\rho^\text{rev} = \psi(\rho(\phi^\text{new}))$.
In this setting, we use $\rho^\text{new}$ as the final new model instead of $\phi^\text{new}$, and our rank merge process employs $\{ \rho^\text{rev}, \phi^\text{old} \}$ and $\{ \rho^\text{new}, \rho^\text{new} \}$ eventually.
%\suseo{Note that this modification only calibrates still does not constrain $\rho^\text{new}$ to be compatible with $\phi^\text{old}$ and preserves its feature discriminability.}
This strategy helps to achieve a better compatibility by allowing both systems to be trainable.
%\suseo{Note that $\rho^\text{new}$ is still not constrained to $\phi^\text{old}$, so its feature

The final loss function to train the reverse transformation module has the identical form to $\mathcal{L}_\text{MCL}$ in \eqref{eq:cmcl} except for the definitions of $s^\text{new}_{ij}$ and $s^\text{old}_{ij}$, which are given by
\begin{align}
s^\text{new}_{ij} &= \exp\left(-\text{dist}\left(\rho^\text{new}(\mathbf{x}_i), \rho^\text{new}(\mathbf{x}_j)\right)\right)  \\ 
s^\text{old}_{ij} &= \exp\left(-\text{dist}\left(\rho^\text{rev}(\mathbf{x}_i), \phi^\text{old}(\mathbf{x}_j)\right)\right). 
\end{align}
This extension still does not constrain $\rho^\text{new}$ to be feature compatible with $\phi^\text{old}$, thus preserving its feature discriminability while further improving distance compatibility.
%Note that this extension does not result in computational overhead at inference stage yet improves the performance even further.

%----------------------------
%          Experiments
% ----------------------------
% !TEX root = ./../main.tex

%%%%%%%%%%%  MAIN TABLE  %%%%%%%%%%%
% !TEX root = ./../../main.tex

\begin{table*}[t]
\begin{center}
\caption{Comparison with existing compatible learning methods on four standard benchmarks in homogeneous model upgrades.
%\suseo{\textit{Gain} denotes the AUC$_\text{mAP}$ ratio of the new model to the old one.}
\textit{Gain} denotes relative gain that each method achieves from old model in terms of AUC$_\text{mAP}$, compared to the gain of new model.
The proposed framework, dubbed as RM, consistently outperforms all other models with significantly large margins for all datasets.
Note that RM$_\text{na\"ive}$ indicates the basic version of distance rank merge described in Sec.~\ref{sec:formulation} and that \textit{Old} and \textit{New} denote embedding models of gallery images.
}
\vspace{-0.1cm}
\label{tab:final_result}
 \scalebox{0.82}{
\setlength\tabcolsep{4pt} \hspace{-0.2cm}
\begin{tabular}{c|cc|c|cc|c|cc|c|cc|c}
\toprule
&\multicolumn{3}{c|}{ImageNet-1K} & \multicolumn{3}{c|}{CIFAR-100} & \multicolumn{3}{c|}{Places-365}  & \multicolumn{3}{c}{Market-1501} \\
& AUC$_\text{mAP}$ & AUC$_\text{CMC}$ & Gain & AUC$_\text{mAP}$ & AUC$_\text{CMC}$ & Gain & AUC$_\text{mAP}$ & AUC$_\text{CMC}$ & Gain & AUC$_\text{mAP}$ & AUC$_\text{CMC}$ & Gain \\
\hline
Old & 31.2 & 49.7 & 0\% & 21.6 & 34.3 & 0\% & 16.5 & 30.7 & 0\% & 62.7 & 82.7 & 0\%  \\
New & 51.3 & 70.3 & 100\% & 47.4 & 62.6 & 100\% & 23.4 & 39.1 & 100\% & 77.3 & 90.9 & 100\%  \\
\hdashline
%Merge [Old$\rightarrow$New] (Ours)
RM$_\text{na\"ive}$ (Ours)      & 40.0 & 63.9 & 44\% & 30.8 & 49.1 & 36\% & 19.5 & 35.8 & 43\% & 69.2 & 87.0 & 45\% \\
\hline
BCT~\cite{bct}                  & 32.0 & 46.3 & 4\% & 26.4 & 43.5 & 19\% & 17.5 & 37.0 & 14\% & 66.6 & 84.3 & 27\%\\ 
FCT~\cite{fct}                  & 36.9 & 58.7 & 28\% & 27.1 & 49.4 & 21\% & 22.5 & 37.3 & 87\% & 66.4 & 84.2 & 25\% \\
FCT (w/ side-info)~\cite{fct}   & 43.6 & 65.0 & 62\% & 37.0 & 53.9 & 60\% & 23.7 & 38.3 & 104\% & 66.4 & 84.4 & 25\%\\
BiCT~\cite{bict}                & 35.1 & 59.7 & 19\% & 29.0 & 48.3 & 29\% & 19.0 & 34.9& 36\% & 65.0 & 82.4 & 16\% \\
\textbf{RM (Ours)}             & \textbf{53.4} & \textbf{68.1} & \textbf{110\%} & \textbf{41.4} & \textbf{60.7} & \textbf{78\%} & \textbf{28.2} & \textbf{41.7} & \textbf{170\%} & \textbf{70.7} & \textbf{87.6} & \textbf{55\%} \\
\bottomrule
\end{tabular}
}
\end{center}
\vspace{-4mm}
\end{table*}

\begin{figure*}
    \centering
    \begin{subfigure}[b]{0.475\textwidth}
        \centering
        \includegraphics[width=\textwidth]{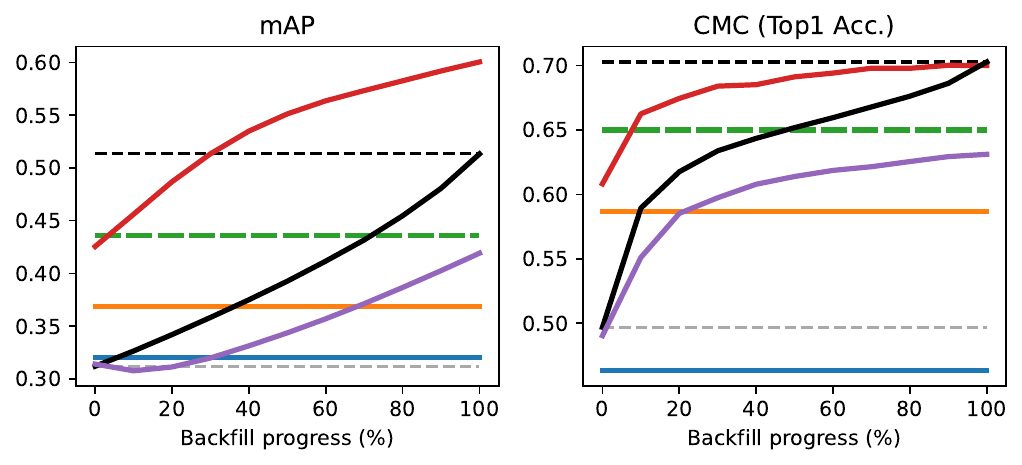}
        \caption[Network2]%
        {{\small ImageNet-1K}}    
        \label{fig:mean and std of net14}
    \end{subfigure}
    \hfill
    \begin{subfigure}[b]{0.475\textwidth}  
        \centering 
        \includegraphics[width=\textwidth]{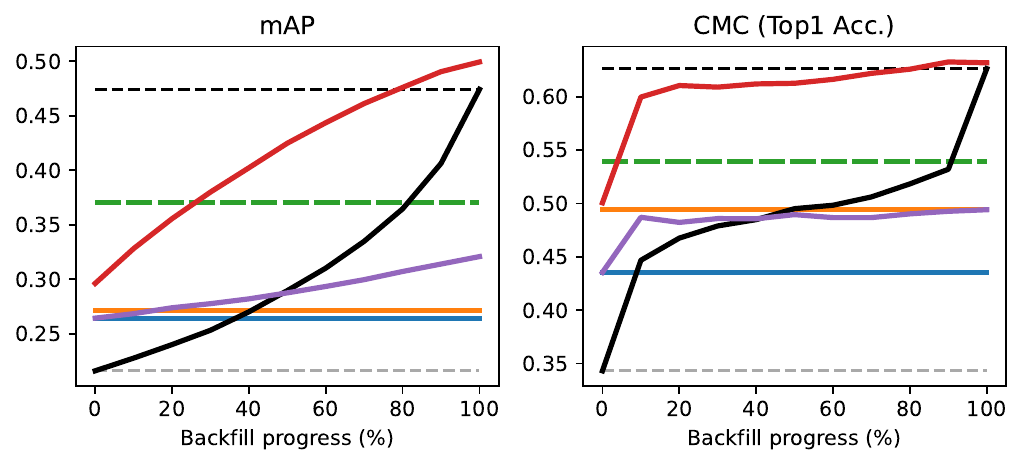}
        \caption[]%
        {{\small CIFAR-100}}    
        \label{fig:mean and std of net24}
    \end{subfigure}

    \begin{subfigure}[b]{0.475\textwidth}  
        \centering 
        \includegraphics[width=\textwidth]{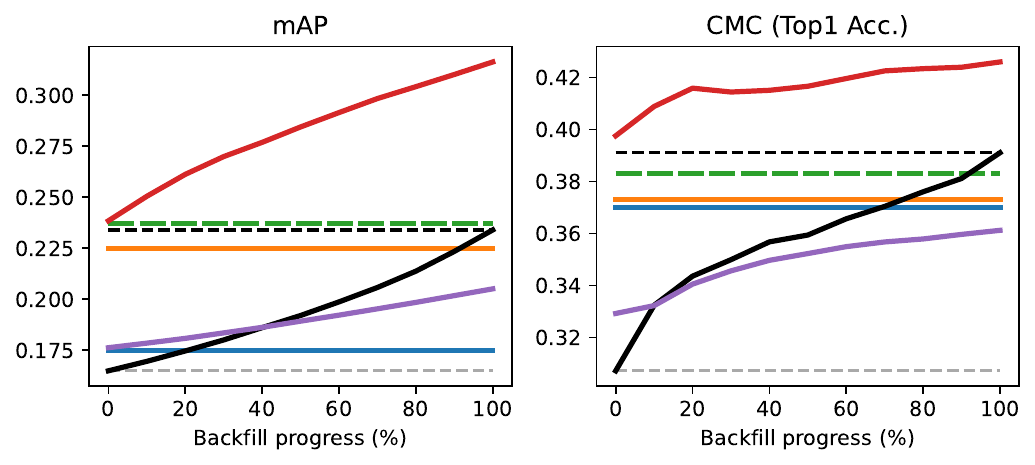}
        \caption[]%
        {{\small Places-365}}    
        \label{fig:mean and std of net24}
    \end{subfigure}
    \hfill
    \begin{subfigure}[b]{0.475\textwidth}  
        \centering 
        \includegraphics[width=\textwidth]{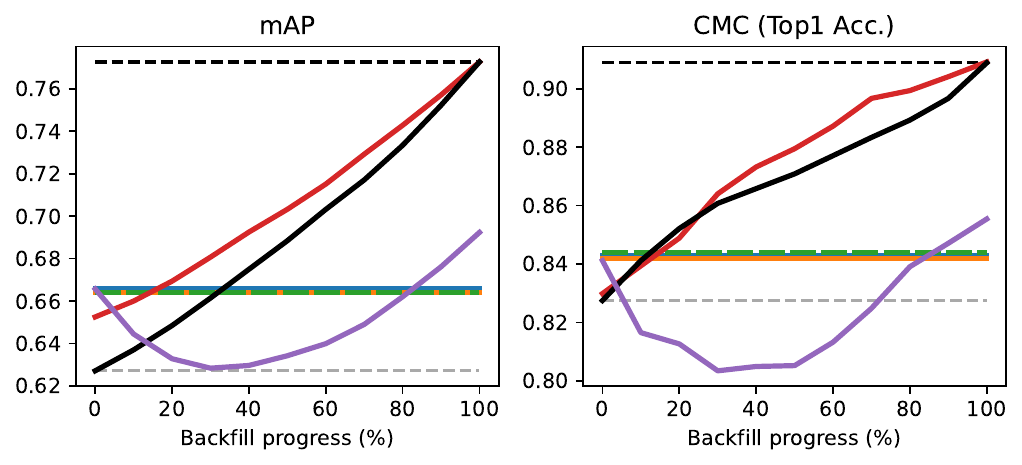}
        \caption[]%
        {{\small Market-1501}}    
        \label{fig:mean and std of net24}
    \end{subfigure}
    \begin{subfigure}[b]{\textwidth}
    \vspace{0.1cm}
        \centering 
        \includegraphics[width=\textwidth]{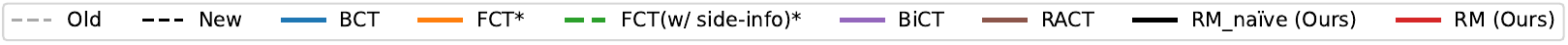}
        \label{fig:mean and std of net24}
        \vspace{-4mm}
    \end{subfigure}
    \caption[ The average and standard deviation of critical parameters ]
    {\small mAP and CMC (Top-1 Acc.) results of our full framework in comparison to existing approaches.
    The numbers in the legend indicate either AUC$_\text{mAP}$ or AUC$_\text{CMC}$ scores.
    % Compared to naive ensemble (green line), ensemble with reverse transformation module (red line) requires only a single feature extraction process but achieves better performance.
    % Although backfilling cost of BiCT (orange line) is cheaper than other ensemble methods, BiCT suffers from limited final performance.
    } 
    \label{fig:final_cifar}
    \vspace{-0.15cm}
\end{figure*}
%%%%%%%%%%%%%%%%%%%%%%%%%%%%%%%%%%%%

\section{Experiments}
\label{sec:experiments}
%We present our experiment setting, the performance of the proposed approach, and results from the analysis of algorithm characteristics. 

\subsection{Dataset and Evaluation Protocol}
\label{sec:setting}

We employ four standard benchmarks, which includes ImageNet-1K~\cite{imagenet}, CIFAR-100~\cite{cifar100}, Places-365~\cite{places365}, Market-1501~\cite{market1501}.
As in previous works~\cite{bct, fct}, we adopt the extended-class setting in model upgrade; the old model is trained with examples from a half of all classes while the new model is trained with all samples.
For example, on the ImageNet-1K dataset, the old model is trained with the first 500 classes and the new model is trained with the whole 1,000 classes.

Following the previous works~\cite{fct, bict, ract}, we measure mean average precision (mAP) and cumulative matching characteristics (CMC)\footnote{CMC corresponds to top-\textit{k} accuracy, and we report top-1 accuracy in all tables and graphs.}.
We also report our comprehensive results in terms of  $\text{AUC}_\text{mAP}$ and $\text{AUC}_\text{CMC}$ at 10 backfill time slices, \ie, $t \in \{0.0, 0.1, ..., 1.0\}$ in~\eqref{eq:rankmerge_metric}.

\subsection{Implementation Details}
We employ ResNet-18~\cite{he2016deep}, ResNet-50~\cite{he2016deep}, and ViT-B/32~\cite{dosovitskiy2020image} as our backbone architectures for either old or new models.
All transformation modules, $\psi(\cdot)$ and $\rho(\cdot)$, consist of 1 to 5 linear layer blocks, where each block is composed of a sequence of operations, (Linear $\rightarrow$ BatchNorm $\rightarrow$ ReLU), except for the last block that only has a Linear layer.
Our algorithm does not use any side-information.
Our modules are trained with the Adam optimizer~\cite{kingma2014adam} for 50 epoch, where the learning rate is $1 \times 10^{-4}$ at the beginning and decayed using cosine annealing~\cite{loshchilov2016sgdr}.
%We use old model's classification score to decide the backfill order for all comparisons.
%Our frameworks are implemented with the Pytorch~\cite{paszke2019pytorch} library and we plan to release the source codes of our work.

\subsection{Results}

\paragraph{Homogeneous model upgrade} We present the quantitative results in the homogeneous model upgrade scenario, where old and new models have the same architecture.
We employ ResNet-50 for ImageNet and ResNet-18 for other datasets.
Table~\ref{tab:final_result} and Figure~\ref{fig:final_cifar} compare the proposed framework, referred to as RM (Rank Merge), with existing compatible learning approaches, including BCT~\cite{bct}, FCT~\cite{fct}, BiCT~\cite{bict}, and RACT~\cite{ract}.
As shown in the table, RM consistently outperforms all the existing compatible training methods by remarkably significant margins in all datasets. 
%\suseo{Because our new feature embedding is learnable under metric compatible training, the final performance of our framework even outperforms the performance of new model in most cases.}
BCT~\cite{bct} learns backward compatible feature representations, which is backfill-free, but its performance gain is not impressive.
FCT~\cite{fct} achieves meaningful performance improvement by transforming old gallery features, but most of the gains come from side-information~\cite{chen2020simple}.
For example, if side-information is not available, the performance gain of FCT drops from 62\% to 28\% on the ImageNet dataset.
Also, such side-information is not useful for the re-identification dataset, Market-1501, mainly because the model for the side-information is trained for image classification using the ImageNet dataset, which shows its limited generalizability.
On the other hand, although BiCT~\cite{bict} takes advantage of online backfilling with less backfilling cost, it suffers from degraded final performance and negative flips in the middle of backfilling.
Note that RM$_\text{na\"ive}$, our na\"ive rank merging between old and new models, is already competitive to other approaches.

\vspace{-2mm}
\paragraph{Heterogeneous model upgrade}

\begin{figure}
    \centering
    
    \begin{subfigure}[b]{0.475\textwidth}  
        \centering 
        \includegraphics[width=\textwidth]{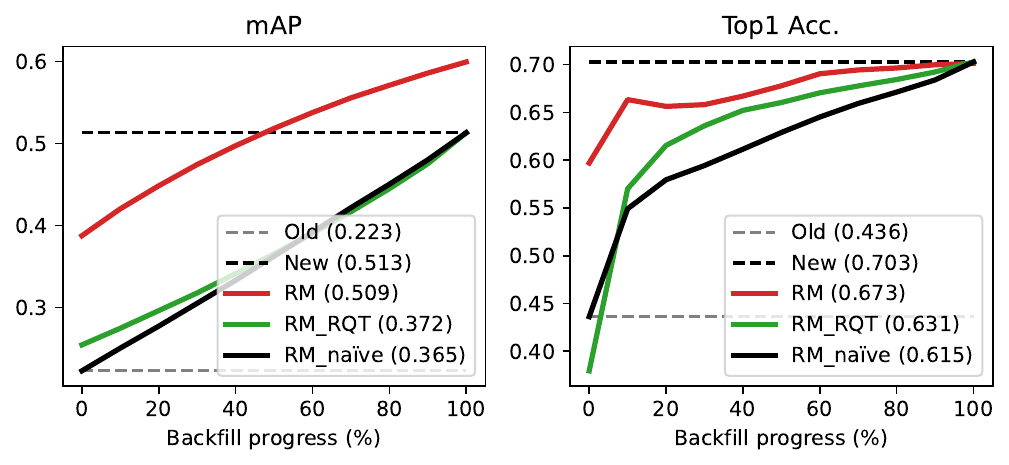}
        \caption[]%
        {{\small ImageNet (ResNet-18 $\rightarrow$ ResNet-50) } }    
        \label{fig:mean and std of net24}
    \end{subfigure}
%    
%    \begin{subfigure}[b]{0.475\textwidth}
%    \centering
%    \includegraphics[width=\textwidth]{figures/exp/hetero/cifar_hetero_res50.pdf}
%    \caption[Network2]%
%    {{\small CIFAR-100 (ResNet-18 $\rightarrow$ ResNet-50) }}    
%    \label{fig:mean and std of net14}
%    \end{subfigure}
    
    \begin{subfigure}[b]{0.475\textwidth}
    \centering
    \includegraphics[width=\textwidth]{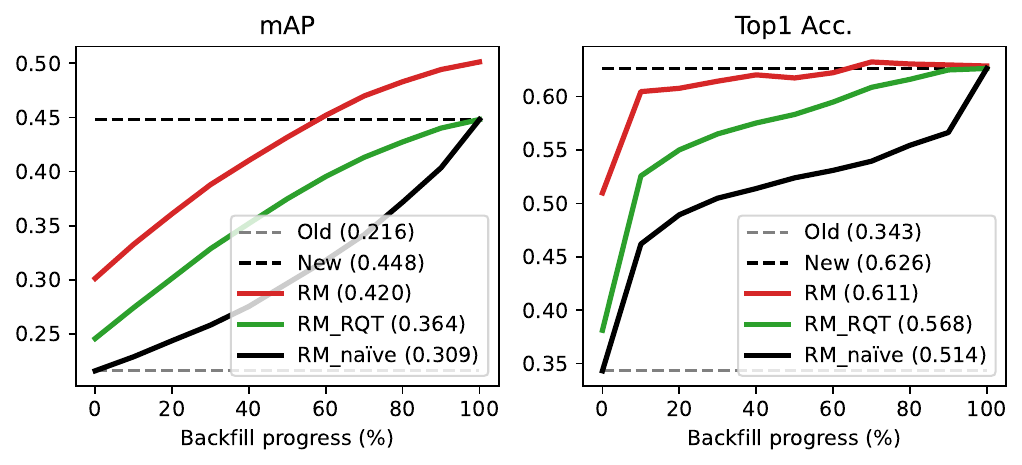}
    \caption[ ]%
    {{\small CIFAR-100 (ResNet-18 $\rightarrow$ ViT-B/32) }}    
    \label{fig:mean and std of net14}
    \end{subfigure}
    
%    \begin{subfigure}[b]{0.475\textwidth}
%    \centering
%    \includegraphics[width=\textwidth]{figures/exp/hetero/market_hetero.pdf}
%    \caption[Network2]%
%    {{\small Market-1501 (ResNet-18 $\rightarrow$ ResNet-50)}}    
%    \label{fig:mean and std of net14}
%    \end{subfigure}
    
%     \begin{subfigure}[b]{0.475\textwidth}
%    \centering
%    \includegraphics[width=\textwidth]{figures/exp/hetero/places_res50.pdf}
%    \caption[Network2]%
%    {{\small Places-365 (ResNet-18 $\rightarrow$ ResNet-50)}}    
%    \label{fig:mean and std of net14}
%    \end{subfigure}
    
    \vspace{-2mm}
    \caption[  ]
    {\small Experimental results with heterogeneous model upgrades.
    Our na\"ive rank merge between different architectures still achieves promising performance curves in various settings, and our full algorithm exhibits significantly better results.
    } 
%    \vspace{-3mm}
    \label{fig:hetero}
\end{figure}

We evaluate our framework in more challenging scenarios and present the results in Figure~\ref{fig:hetero}, where the old and new models have different architectures, \eg, ResNet-18 $\rightarrow$ ResNet-50 or ResNet-18 $\rightarrow$ ViT-B/32.
In this figure, RM$_\text{RQT}$ (green line) denotes our ablative model trained with~\eqref{eq:rev_basic}.
Even in this setting, where both embedding spaces are more incompatible, our rank merge results from the old and new models still manage to achieve a monotonous performance growth curve and RM improves the overall performance significantly further, which validates the robustness of our frameworks.
Please refer to Table~\ref{tab:final_result_hetero} for additional results.

\vspace{-2mm}
\paragraph{Ablation study}

\begin{figure}
    \centering

    \begin{subfigure}[b]{0.475\textwidth}  
        \centering 
        \includegraphics[width=\textwidth]{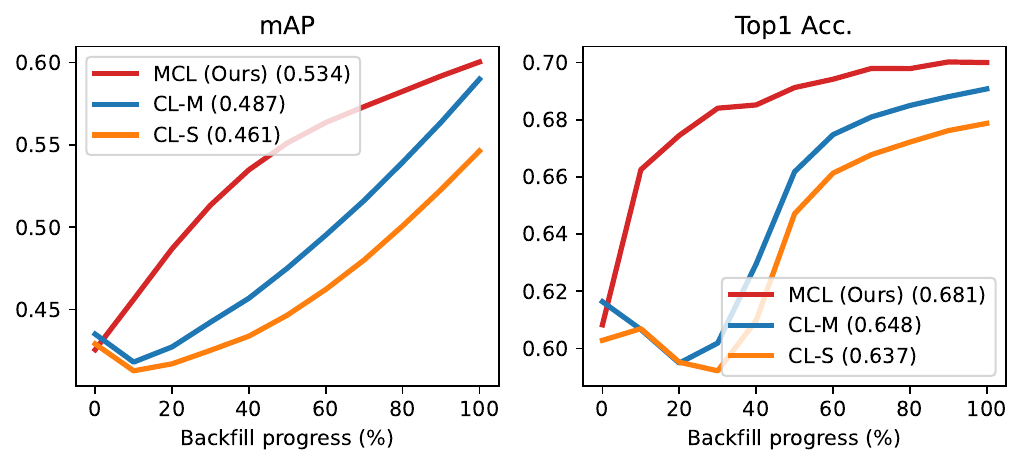}
        \caption[]%
        {{\small ImageNet (ResNet-50 $\rightarrow$ ResNet-50) }}    
        \label{fig:image_mcl}
    \end{subfigure}

%    \begin{subfigure}[b]{0.475\textwidth}
%    \centering
%        \includegraphics[width=\textwidth]{figures/exp/abl/cifar_vit.pdf}
%    \caption[Network2]%
%    {{\small CIFAR-100 (ViT-B/32 $\rightarrow$ ViT-B/32)  }}    
%    \label{fig:cifar_mcl}
%    \end{subfigure}
        
    \begin{subfigure}[b]{0.475\textwidth}
    \centering
    \includegraphics[width=\textwidth]{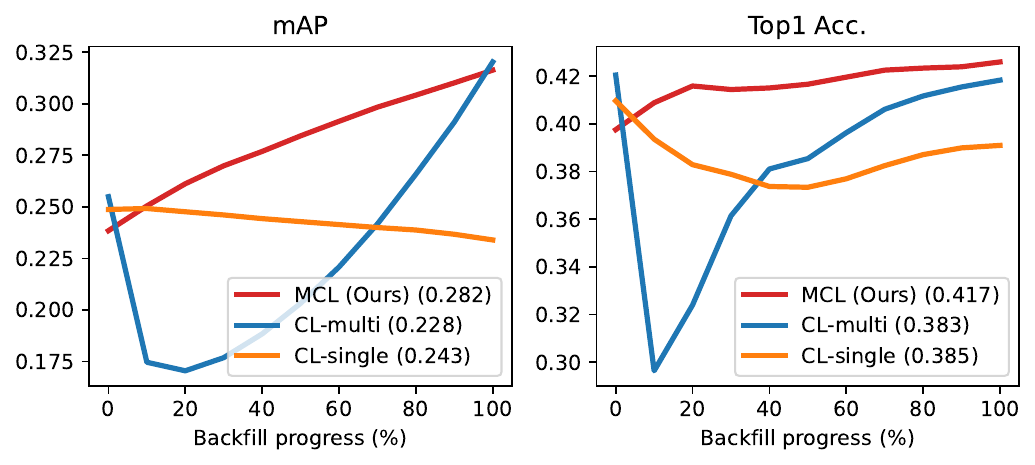}
    \caption[Network2]%
    {{\small Places-365 (ResNet-18 $\rightarrow$ ResNet-18)   }}    
    \label{fig:cifar_mcl}
    \end{subfigure}
    
    \vspace{-2mm}    
    \caption[ The average and standard deviation of critical parameters ]
    {\small Ablation study of the metric compatible contrastive learning loss on several datasets, which validates that the distance calibration plays a crucial role for effective rank merging.} 
%        CMCL outperforms other ablative models, CL-M and CL-S, 
    \label{fig:abl_ccl}
%    \vspace{-2mm}
\end{figure}

We analyze the results from the ablations of models for our metric compatible contrastive learning.
For compatible training, CL-S employs contrastive learning within the backward system only as in~\eqref{eq:cl} while our MCL considers distance metrics from both backward and new retrieval systems simultaneously as in~\eqref{eq:cmcl}.
%For a more thorough ablation study, we also design and test another metric learning objective, called CL-M, which is given by
%%
%\begin{align}
%\hspace{-2mm} \mathcal{L}_\text{CL-M}(\mathbf{x}_i, y_i) = & -\log \frac{\sum_{y_k=y_i} s_{ik}^\text{old}}{\sum_{y_k=y_i} s_{ik}^\text{old} + \sum_{y_k\neq y_i} s_{ik}^\text{old}} 
%\nonumber \\
%& -\log \frac{\sum_{y_k=y_i} s_{ik}^\text{new}}{\sum_{y_k=y_i} s_{ik}^\text{new} + \sum_{y_k\neq y_i} s_{ik}^\text{new}},
%\label{eq:cl2}
%\end{align}
%%
%which conducts contrastive learning for both backward and new retrieval systems separately.
For a more thorough ablation study, we also design and test another metric learning objective, called CL-M, which conducts contrastive learning for both backward and new retrieval systems separately as
\begin{align}
\hspace{-2mm} \mathcal{L}_\text{CL-M}(\mathbf{x}_i, y_i) = & -\log \frac{\sum_{y_k=y_i} s_{ik}^\text{old}}{\sum_{y_k=y_i} s_{ik}^\text{old} + \sum_{y_k\neq y_i} s_{ik}^\text{old}} 
\nonumber \\
& -\log \frac{\sum_{y_k=y_i} s_{ik}^\text{new}}{\sum_{y_k=y_i} s_{ik}^\text{new} + \sum_{y_k\neq y_i} s_{ik}^\text{new}}.
\label{eq:cl2}
\end{align}
Figure~\ref{fig:abl_ccl} visualizes the results from the ablation studies, where MCL consistently outperforms both CL-S and CL-M in various datasets and architectures.
CL-M generally gives better merge results than CL-S because it calibrates the distances of new retrieval system additionally.
However, CL-M still suffers from negative flips because the distance metrics of both retrieval systems are calibrated independently and not learned to be directly comparable to each other.
On the other hand, MCL improves overall performance curves consistently without negative flips.
This validates that considering the distance metrics of both systems simultaneously helps to achieve better metric compatibility and consequently stronger merge results.

\begin{figure}
    \centering
        \hspace{-0.1cm}
    \begin{subfigure}[b]{0.236\textwidth}  
        \centering 
        \includegraphics[width=\textwidth]{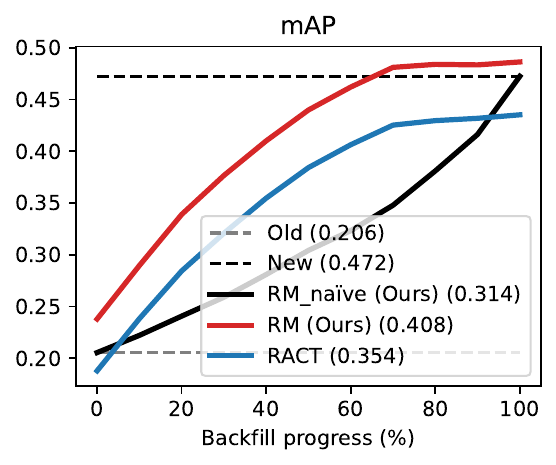}
        \caption[]%
        {{\small CIFAR-100 } }    
        \label{fig:mean and std of net24}
    \end{subfigure}
    \begin{subfigure}[b]{0.236\textwidth}  
        \centering 
        \includegraphics[width=\textwidth]{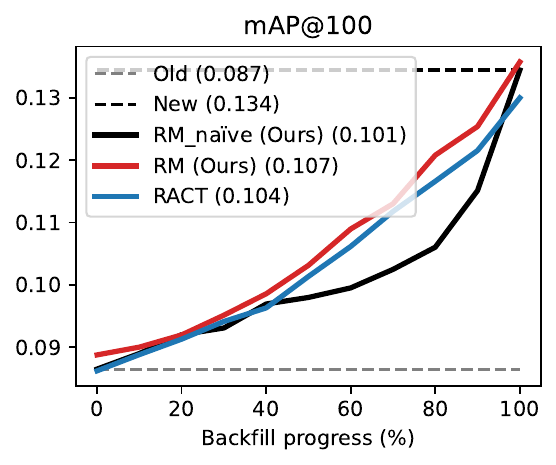}
        \caption[]%
        {{\small GLDv2-test } }    
        \label{fig:mean and std of net24}
    \end{subfigure}  
    \vspace{-5mm}
    \caption[  ]
    {\small Experimental results in open-class setting.
Our framework does not sacrifice the final performance of new model and achieves superior merge results during backfilling.
%    Our na\"ive rank merge between different architectures still achieves promising performance curves in various settings, and our full algorithm exhibits significantly better results.
    } 
%    \vspace{-2mm}
    \label{fig:open_class}
\end{figure}

%\vspace{-3mm}
\paragraph{Open-class setting}
We apply the proposed approach to an open-class setting, where the old and new data do not share the same classes, on CIFAR-100 and Google Landmarks V2 (GLDv2)~\cite{gldv2} with the ResNet-18 backbone network.
For both datasets, we randomly sample 30\% of classes for training the old model while the rest of the classes are used for training the new model.
%Although RACT achieves backward compatibility, it sacrifices the discriminability of new model.
%[Add discussion about RM!!]
%As presented in Figure~\ref{fig:open_class}, RM$_\text{na\"ive}$ still works well without any algorithm changes on both datasets under the open-class setting.
In this setting, RM consistently outperforms RACT during backfilling on both datasets as presented in Figure~\ref{fig:open_class}.
%As presented in Figure~\ref{fig:open_class}, RM$_\text{na\"ive}$ still achieves monotonic performance improvement from old to new models while RM improves the overall performance even further on both datasets under the open-class setting.
%RACT achieves backward compatibility but it inevitably sacrifice the feature discriminability of new model.
Although RACT achieves meaningful performance gains in the middle of backfilling, it sacrifices the accuracy after the completion of backfilling at the expense of achieving feature compatibility.
To the contrary, RM is free from the issue and even outperforms the original new model.
This fact validates that our metric compatible training preserves feature discriminability while properly calibrating the scales of distance metrics between the old and new systems.
RM$_\text{na\"ive}$ also improves accuracy monotonically from old to new models with no extra training. 
%Again, our RM$_\text{na\"ive}$ works well without any extra training in the open-class setting consistently.

%----------------------------
%          Conclusion
% ----------------------------
% !TEX root = ./../main.tex

\section{Conclusion}

\label{sec:conclusion}

We presented a novel compatible training framework for effective and efficient online backfilling.
We first addressed the inherent trade-off between compatibility and discriminability, and proposed a practical alternative, online backfilling, to handle this dilemma.
Our distance rank merge framework elegantly sidesteps this issue by bridging the gap between old and new models, and our metric-compatible learning further enhances the merge results with distance calibration.
Our framework was validated via extensive experiments with significant improvement.
We believe our work will provide a fundamental and practical foundation for promoting new directions in this line of research.

\paragraph{Acknowledgements}
This work was partly supported by the National Research Foundation of Korea grant [No.2022R1A2C3012210] and the Institute of Information communications Technology Planning \& Evaluation (IITP) grants [No.RS-2022-II220959, No.RS-2021-II211343, No.RS-2021-II212068], and the National Research Foundation of Korea (NRF) grant [No.RS-2022-NR070855], funded by the Korean government (MSIT).

%%%%%%%%% REFERENCES
{\small
\bibliographystyle{ieee_fullname}
\bibliography{egbib}
}

% !TEX root = ./main.tex

\newpage
\appendix
\onecolumn

\renewcommand{\thesection}{\Alph{section}} 
\renewcommand{\thetable}{A\arabic{table}}
\renewcommand{\thefigure}{A\arabic{figure}}

\clearpage

\section{Additional Experimental Results}

%This section presents additional experimental results that were not included in the main paper due to space limitations.

\label{sec:add_exp}

%\paragraph{Full Experimental Results with Heterogenous Model Upgrades}
\paragraph{Full experimental results with heterogenous model upgrades}

\begin{table*}[h!]
\begin{center}
\caption{Experimental results with heterogeneous model upgrades, from ResNet-18 (old) to ResNet-50 (new), on four standard benchmarks.
\textit{Gain} denotes relative gain that each method achieves from old model in terms of AUC$_\text{mAP}$, compared to the gain of new model.
Our full algorithm (RM) outperforms all other existing approaches with significant margins.
}
\vspace{-0.2cm}
\label{tab:final_result_hetero}
 \scalebox{0.85}{
\setlength\tabcolsep{4pt} \hspace{-0.2cm}
\begin{tabular}{c|cc|c|cc|c|cc|c|cc|c}
\toprule
&\multicolumn{3}{c|}{ImageNet-1K} & \multicolumn{3}{c|}{CIFAR-100} & \multicolumn{3}{c|}{Places-365}  & \multicolumn{3}{c}{Market-1501} \\
& AUmAP & AuCMC & Gain & AUmAP & AuCMC & Gain & AUmAP & AuCMC & Gain & AUmAP & AuCMC & Gain \\
\hline
Old & 22.3 & 43.6 & 0\% & 21.6 & 34.3 & 0\% & 16.5 & 30.7 & 0\% & 62.7 & 82.7 & 0\%  \\
New & 51.3 & 70.3 & 100\% & 50.0 & 67.5 & 100\% & 23.4 & 39.4 & 100\% & 82.1 & 92.9 & 100\%  \\
\hdashline
RM$_\text{na\"ive}$ (Ours)     & 36.5 & 61.5 & 49\% & 33.8 & 48.9 & 43\% & 20.3 & 35.6 & 55\% & 72.2 & 82.3 & 49\% \\
\hline
BCT~\cite{bct}                  & 30.4 & 53.3 & 28\% & 27.1 & 45.9 & 19\% & 18.6 & 34.2 & 30\% & 66.6 & 85.0 & 20\% \\ 
FCT~\cite{fct}                  & 43.2 & 63.6 & 72\% & 28.9 & 54.2 & 26\% & 23.7 & 38.5 & 104\% & 67.6 &  85.3 & 25\% \\
FCT (w/ side-info)~\cite{fct}   & 47.5 & 65.0 & 87\% & 41.5 & 64.2 & 70\% & 24.9 & 39.5 & 122\% & 68.0 & 85.1 & 27\%\\
BiCT~\cite{bict}                & 33.8 & 58.8 & 40\% & 29.9 & 50.7 & 29\% & 21.4 & 36.7 & 71\% &65.1 & 84.4& 12\% \\
RACT~\cite{ract} 		& 38.9 & 55.1 & 57\% & 36.9& 60.9 & 54\% & 22.8& 39.3& 89\% & 73.0 & 89.9 & 53\% \\
\textbf{RM (Ours)}             & \textbf{50.9} & \textbf{67.3} & \textbf{99\%} & \textbf{45.6} & \textbf{64.5} & \textbf{85\%} & \textbf{29.2} & \textbf{41.8} & \textbf{184\%} & \textbf{74.7} & \textbf{90.0} & \textbf{62\%} \\
\bottomrule
\end{tabular}
 }
\end{center}
\vspace{-0.4cm}
\end{table*}

\begin{figure*}[h]
    \centering
    \begin{subfigure}[b]{0.475\textwidth}
        \centering
        \includegraphics[width=\textwidth]{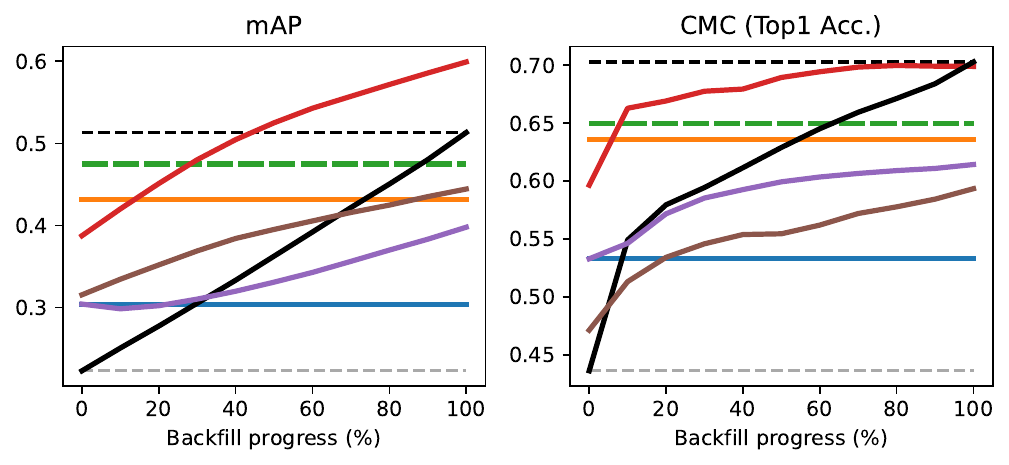}
        \caption[Network2]%
        {{\small ImageNet-1K}}    
        \label{fig:mean and std of net14}
    \end{subfigure}
    \hfill
    \begin{subfigure}[b]{0.475\textwidth}  
        \centering 
        \includegraphics[width=\textwidth]{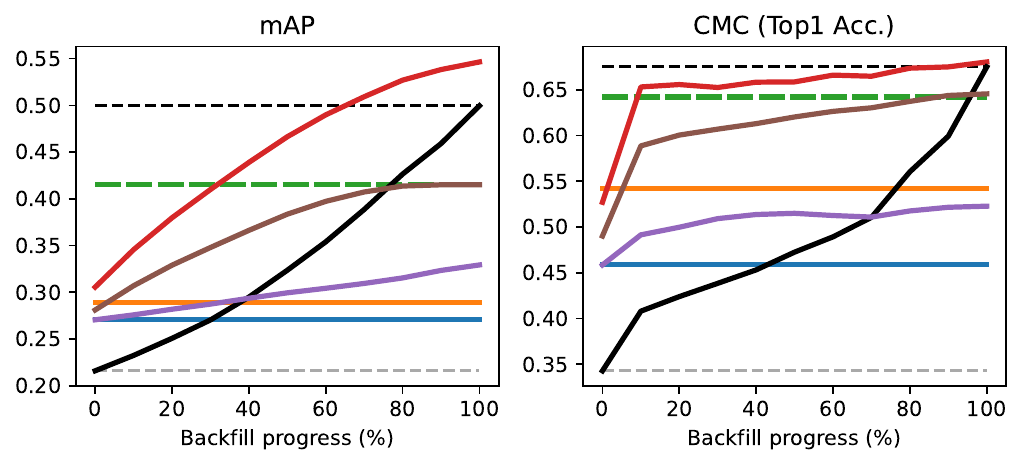}
        \caption[]%
        {{\small CIFAR-100}}    
        \label{fig:mean and std of net24}
    \end{subfigure}

    \begin{subfigure}[b]{0.475\textwidth}  
        \centering 
        \includegraphics[width=\textwidth]{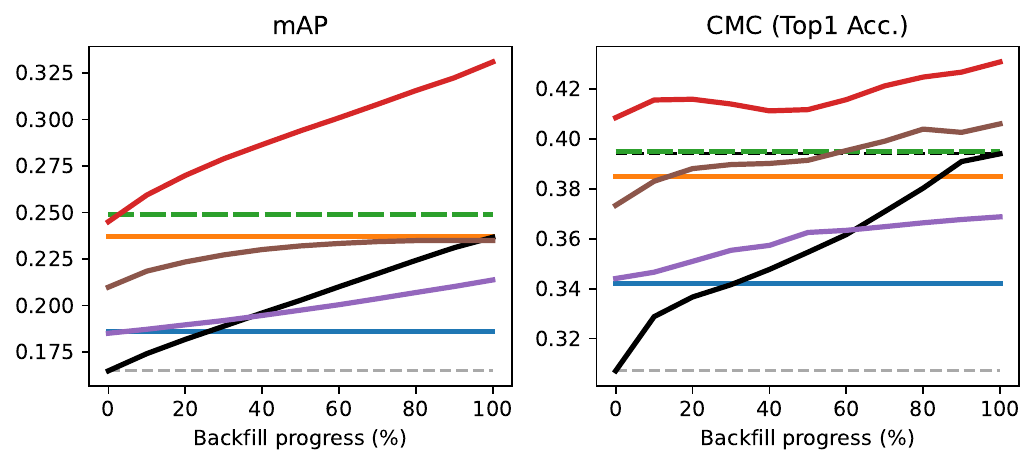}
        \caption[]%
        {{\small Places-365}}    
        \label{fig:mean and std of net24}
    \end{subfigure}
    \hfill
    \begin{subfigure}[b]{0.475\textwidth}  
        \centering 
        \includegraphics[width=\textwidth]{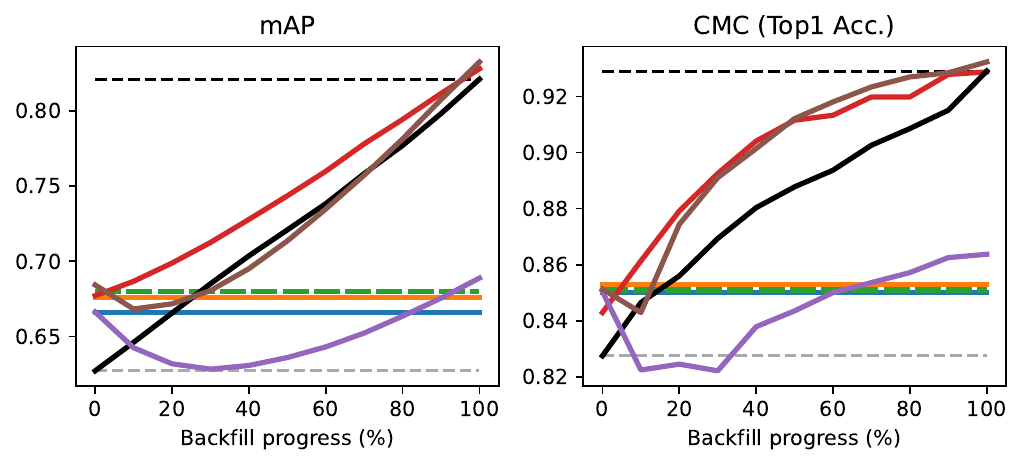}
        \caption[]%
        {{\small Market-1501}}    
        \label{fig:mean and std of net24}
    \end{subfigure}
    \begin{subfigure}[b]{\textwidth}
    \vspace{0.2cm}
        \centering 
        \includegraphics[width=\textwidth]{./figures/exp/final/final_legend.pdf}
        \label{fig:mean and std of net24}
        \vspace{-0.2cm}
    \end{subfigure}
    \caption[ The average and standard deviation of critical parameters ]
    {\small mAP and CMC (Top-1 Acc.) results of our algorithms in comparison to existing approaches under heterogenous model upgrades.
    The numbers in the legend indicate either AUC$_\text{mAP}$ or AUC$_\text{CMC}$ scores.
    % Compared to naive ensemble (green line), ensemble with reverse transformation module (red line) requires only a single feature extraction process but achieves better performance.
    % Although backfilling cost of BiCT (orange line) is cheaper than other ensemble methods, BiCT suffers from limited final performance.
    } 
    \label{fig:final_curve_hetero}
\end{figure*}

%\paragraph{Full results with heterogeneous model upgrade}
Table~\ref{tab:final_result_hetero} and Figure~\ref{fig:final_curve_hetero} present the full experimental results in comparison to existing compatible learning approaches~\cite{bct, fct, bict, ract} in a heterogeneous model upgrade scenario, which supplements Figure 8 of the main paper. 
In this scenario, \textit{Old} and \textit{New} models employ ResNet-18 and ResNet-50 architectures, respectively.
%As in the homogeneous model upgrade (Table 1 and Figure 7 of the main paper), our na\"ive distance rank merge framework between old and new models still provides monotonically increasing results between different backbone architectures consistently.
As in the homogeneous model upgrade (Table 1 and Figure 7 of the main paper), our na\"ive distance rank merge framework between old and new models still provides monotonically increasing results throughout the backfilling process.
Our final algorithm, dubbed as RM, significantly outperforms all other methods on all datasets, which validates its excellence in a more challenging scenario.

\paragraph{Backfilling strategy}
\label{sec:backfill}
Another possible improvement we can make is by selecting the right samples to be backfilled first. 
With the results so far, even random sample selection is sufficient to achieve strong and robust retrieval merge, but there may be room for further improvement.
If we select less reliable samples to be backfilled first among gallery images, then those will benefit more than other samples from the model upgrade, improving the performance at the same backfilling cost.

We introduce two ways to measure the confidence of each old gallery sample: classification score and cosine similarity from centroid.
The former uses the old classifier's final score, which reveals samples that have a lower confidence among the old gallery samples.
However, it requires the classification layer of old model, which may not be applicable in some training regimes like as self-supervised learning.
For the latter, we first get classwise centroid embedding by calculating the average of feature embeddings for each class, and measure the cosine similarity between each sample and its class's centroid.
Low scoring samples are backfilled first based on one of these measurements, and we adopt the former in our experiments.

%\paragraph{Additional Results}

\begin{figure}
    \centering
    % \hfill
    \begin{subfigure}[b]{0.27\linewidth}  
        \centering 
        \includegraphics[width=\textwidth]{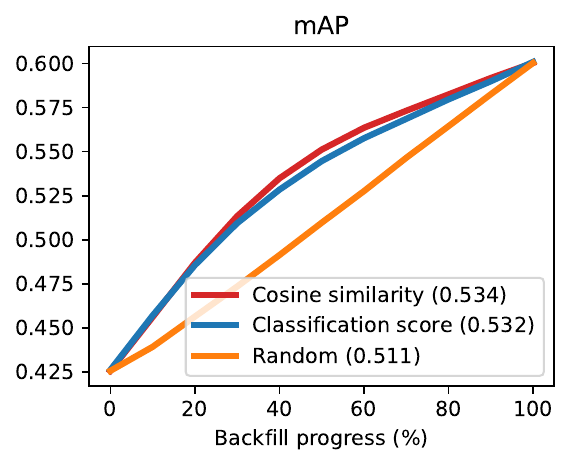}
        \caption[]%
        {{\small ImageNet-1K}}    
        \label{fig:mean and std of net24}
    \end{subfigure}
    \hspace{0.3cm}
        \begin{subfigure}[b]{0.27\linewidth}
        \centering
        \includegraphics[width=\textwidth]{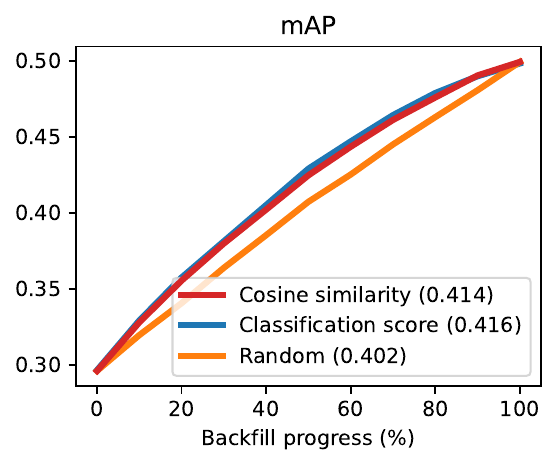}
        \caption[Network2]%
        {{\small CIFAR-100}}    
        \label{fig:mean and std of net14}
    \end{subfigure}
        \hspace{0.3cm}
     \begin{subfigure}[b]{0.27\linewidth}
        \centering
        \includegraphics[width=\textwidth]{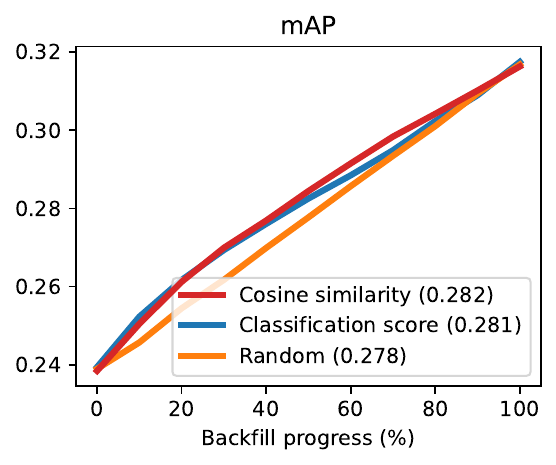}
        \caption[Network2]%
        {{\small Places-365}}    
        \label{fig:mean and std of net14}
    \end{subfigure}
    \caption[Backfilling strategy]
    {\small Ablation study of the backfilling order on ImageNet-1K, CIFAR-100, and Places-365 datasets.
    To determine the backfilling order, compared to random selection, adopting either cosine similarity or classification score gives better performance at the same backfilling cost.
%    If uncertain samples are backfilled first, we can achieve better performance at the same backfilling cost, compared to random selection.
    % Compared to naive ensemble (green line), ensemble with reverse transformation module (red line) requires only a single feature extraction process but achieves better performance.
    % Although backfilling cost of BiCT (orange line) is cheaper than other ensemble methods, BiCT suffers from limited final performance.
    } 
    \label{fig:ablation_backfill}
\end{figure}

We conduct ablative study on the backfilling order with different strategies using our framework in Figure~\ref{fig:ablation_backfill}.
`Cosine similarity' indicates that the ordering is sorted by the cosine similarity between each sample and its class's centroid, while `Classification score' denotes the order is based on the final score of classification layer in old model.
In each case, samples with low scores are backfilled first.
Compared to random selection, both strategies improves the average performance during backfilling, while the gain is more significant in the large-scale dataset, ImageNet.

%\paragraph{Backfilling strategy}
%\fi

%\paragraph{Calibration and feature discriminability}
%We found that calibration improves.

\section{Experimental Details}

\subsection{Datasets}
\paragraph{ImageNet-1K}\hspace{-0.2cm}\cite{imagenet} is a large-scale image recognition dataset introduced in ILSVRC 2012 challenge, which contains 1000 image classes with 1.2M of training images.
% 50K images for train and validation split, respectively.
The evaluation set has 50,000 images, 50 images per each class.
Following previous works, we use the first 500 classes to train the old model, while the whole classes is used to train the new model.
For evaluation, we employ the full validation split for both query and gallery sets.

\paragraph{CIFAR-100}\hspace{-0.2cm}\cite{cifar100} consists of 100 classes with 50K of training and 10K of validation images. 
The old model is trained with the images of the first 50 classes, and the new model utilizes the whole classes.
We use the entire set of validation images as query and gallery sets for evaluation, as in ImageNet-1K.

\paragraph{Places-365}\hspace{-0.2cm}\cite{places365} is a large-scale scene recognition dataset, which contains 1.8M of training images with 365 scene categories.
We use the first 182 categories to train the old model, while use the whole images to train the new model.
The evaluation set contains 36,500 images, 100 images per each category, all of which are used for evaluation.

\paragraph{Market-1501}\hspace{-0.2cm}\cite{market1501} is a re-identification dataset, which consists of training, gallery, and query splits, with a total of 32,886 images and 1,501 classes.
The training split has 751 classes with 12,936 images.
Among them, the first 375 classes are used to train the old model, while the whole 751 classes are used to train the new model.
For evaluation, we follow the standard re-identification testing process with the pre-defined query and gallery splits, each of which contains 3,368 and 15,913 images, respectively.
% training, query, and gallery splits.
% The training split has 12,936 images with 751 
%The training split has XX images with XX classes.
%Query and gallery splits are for evaluation, where each consists of X.
%CIFAR-100~\cite{cifar100}, Places-365~\cite{places365}, and Market-1501~\cite{market1501}

\paragraph{Google Landmarks V2 (GLDv2)}\hspace{-0.2cm}\cite{gldv2} is a large-scale landmark retrieval dataset, which consists of 1.5M of training images with 81,313 classes.
The test set contains 750 query images and 760K gallery images. 
We randomly sample 30\% of classes for training the old model while the rest of the classes are used for training the new model.
We adopted mAP@100 as an evaluation metric following prior works.

\subsection{Setup}
We adopt the cosine distance as the distance metric for training and evaluation.
The input image is resized to $224\times 224$ for both training and evaluation on all datasets.
In our frameworks, all transformation modules, $\psi(\cdot)$ and $\rho(\cdot)$, consist of 2 lightweight network blocks for (CIFAR-100, Market-1501, GLDv2) and 5 blocks for (ImageNet-1K, Places-365), where each block is composed of a sequence of operations, (Linear $\rightarrow$ BatchNorm $\rightarrow$ ReLU), except for the last block that only has a linear layer.
Following the previous work~\cite{fct}, we set the feature dimension to 512 for the Places-365 dataset and 128 for the others.
For a fair comparison, we decide the backfilling order based on the score of old classifier for all online backfilling methods.
We reproduced BCT, FCT, BiCT, and RACT based on their official codes\footnote{\url{https://github.com/apple/ml-fct}}$^,$\footnote{\url{https://github.com/YantaoShen/openBCT}}$^,$\footnote{\url{https://github.com/TencentARC/OpenCompatible}},  and papers~\cite{bct, fct, bict}.
All our experiments are conducted with 4 NVIDIA A100 GPUs.
%The cost of BiCT; still okay for BiCT-final.

\subsection{Cost Analysis}
In Section 4 of the main paper, our framework introduces lightweight transformation modules for metric compatible training.
Given that the backbone network is ResNet-18, which takes 1.8G of multiply-accumulate operations (MACs), the computational overhead for the transformation modules at the inference stage is only 132.8K of MACs, which is negligible ($<0.01\%$).

\iffalse
\section{Limitation}
Our frameworks provide the generic framework that enables online backfilling in arbitrary model upgrade scenarios. 
One limitation is that we need to access the raw old gallery images for feature re-extraction, which may not be fully available in some practical cases.
We believe that preserving the final performance without accessing raw old gallery images could be a promising future research direction.
\fi

\end{document}